\documentclass[12pt]{article}
\usepackage{graphicx}

\def\Title#1{\begin{center} {\Large #1 } \end{center}}
\def\Author#1{\begin{center}{ \sc #1} \end{center}}
\def\Address#1{\begin{center}{  #1} \end{center}}

\usepackage{graphicx}
\usepackage{multirow}
\usepackage{mathtools}

\begin{document}

\vfill
\Title{Opinion Mining on Non-English Short Text}
\Author{Esra Akbas}
\Address{Florida State University\\ akbas@cs.fsu.ede}
\begin{abstract}
 As the type and the number of such
venues increase, automated analysis of sentiment on textual
resources has become an essential data mining task. In this paper, we investigate the problem of mining opinions on the collection of informal short texts. Both positive and negative sentiment strength of texts are detected. We focus on a non-English language that has few resources for text mining. This approach would help enhance the sentiment analysis in languages where a list of opinionated words does not exist. We propose a new method projects the text into dense and low dimensional feature vectors according to the sentiment strength of the words. We detect the mixture of positive and negative sentiments on a multi-variant scale. Empirical evaluation of the proposed framework on Turkish tweets shows that our approach gets good results for  opinion mining.\\

\end{abstract}

\section{Introduction}
\begin{figure}[h!]
\centering
\includegraphics[scale=0.7] {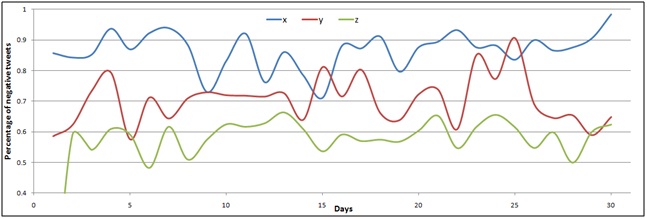}
\caption{\small Sentiment of tweets over time for the three major
wireless carriers 
\label{fig:stot}}
\end{figure}

 Users generally do an online search about a product before buying it, and online reviews about the product affect their opinion significantly. Companies can monitor their brand reputations, analyze how people's opinions change over time, and decide whether a marketing campaign is effective or not. Companies can monitor their brand reputations, analyze how people's opinions change over time, and
 decide whether a marketing campaign is effective or not. Figure \ref{fig:stot} shows a simple example that illustrates the sentiment of tweets over
 time for the three major wireless carriers in Turkey, called Carrier-X, Carrier-Y, and Carrier-Z in this
 paper. There is a clear difference in the overall reputation of these companies on social media. The companies seek to understand what
particular aspects contribute to these differences.

 As sharing opinions increase, getting useful information of sentiments on textual resources has become an essential data mining goal. The task of sentiment
classification aims to identify the polarity of sentiment in text. The
polarity is predicted on either a binary (positive, negative) or a
multi-variant scale as the strength of sentiment expressed in the
text. On the other hand, a text may contain a mix of
 positive and negative sentiment; hence it is often necessary to detect both of them simultaneously \cite{2}. An example of a text that shows this mixture is: ``Carrier-X kaliteli ama \c{c}ok pahal{\i}". It is stated in the example that ``Carrier-X provides quality service, but its cost is too high".

Extracting polarity from textual resources involves many challenges.
Sentiment may be expressed in a more hidden attitude. While a sentence with opinion words does not indicate sentiment, a sentence without any opinion words may contain
sentiment. 
The major problem we face in this paper is to classify aspect-based sentiments of informal short texts in a non-English language.
There have been several studies focusing on sentiment classification on reviews, blogs, and news which are relatively well-formed, coherent
and at least paragraph-length pieces of text. On the other hand,
short-texts include only a few sentences of phrases. When we use all words as
features, corresponding feature vectors are so sparse and we need to select
those that have an important effect on the sentiment of the text. The
other challenge is that most of the available resources, such as lexicons and labeled corpus, in the literature, are in English.
Constructing these resources manually is time-consuming.

Most of the previous studies are in the English text that has a rich resource for sentiment analysis. Also, they use all words in the text to detect sentiment. However, all words may not have an effect on the sentiment. To solve these problems, we propose several methods.

Contributions of this paper are summarized as follows:

\begin{enumerate}

 \item We focus on non-English texts in Twitter that contain informal short messages and construct our resources, lexicon and corpus for Turkish. Our work offers a methodology to enhance the sentiment analysis to other languages where such rich sources do not exist.
\item We present a new method to automatically construct a list of words with their sentiment strengths.
\item We propose a new representation of the text according to sentiment strength of the words, called  \textit{Opinion-based text representation:Grouped} as a new feature vector type. This method projects the text into dense and low dimensional feature vectors in contrast to BofW text representation as sparse and high dimensional.
\item We perform extensive evaluation of our proposed approaches by using our Turkish tweets dataset. According to this evaluation, our approaches give good accuracy results for sentiment analysis.
\end{enumerate}

\section{Background}

\subsection{Sentiment Analysis}

A large collection of research on mining opinions from text has been
done. The existing works are presented as comprehensive survey in \cite{BobLiu12,PANGLEE}. Most of these works detect sentiment as positive-negative, or
add a natural class to them. There are also some studies
\cite{2,TSS}, that detect positive-negative sentiment strength by
predicting human ratings on a scale.

Some approaches use unsupervised (lexical) methods which  utilize a dictionary or a lexicon of pre-tagged (positive-negative or strength of sentiment) opinion words known as polar words and sentiment words \cite{9,2,11}. In addition to manually constructing this word list, it can be constructed automatically \cite{JK04,WPDMLA}.

As a classification problem \cite{28,16},a classifier is trained with a collection of tagged corpus using a chosen feature vector and a supervised machine learning method.  Different kinds of feature vectors can be used to represent text in classification. In addition to words,  tags, emoticons, punctuation \(e.g., ?, !,..\), negation words are also exploited as a feature type \cite{18,19,7,PANGLEE}. 

While most of research are for the text written in English, in recent years, some research is done for text written in
Turkish \cite{TALS,Oddu,TSA,parlar2016new,TSS}. 

\section{Methodology}
In this section, our aspect-based sentiment model is described. First, we have a preprocessing step before converting text into feature vectors. The second step, the sentiment of the tweets are extracted, and finally, the average sentiment value of tweets is computed.  Lexical and supervised approaches are combined to to extract sentiment value of each tweet. In the lexical approach, we need a sentiment word list. Constructing a sentiment word list manually is time consuming. Therefore, an automated method is proposed for this process. Then, we propose two methods to measure the sentiment strength of tweets. In addition to them, SentiStrength\cite{2} is configured for Turkish.

\subsection{Preprocessing}
There is a lot of unnecessary information for sentiment analysis in a text. Before converting the text into feature vectors, we remove these redundant parts of the text by applying different processes, such as punctuation and suffixes removing, spelling correction. A Turkish morphological analyzer, Zemberek \cite{Zemberek}, is used for these processes. Details of the preprocessing steps are skipped because of size limitation.

As another preprocessing step, we study on emoticons, which are also important symbols to represent the opinion in the text. There are different types of emoticons used in  informal text. While some of them are used to express positive emotion, some of them are used for negative emotion. We group emoticons according to their corresponding sentiment value as ``positive", ``negative" and ``others" and a symbol is given to each group. Then, all emoticons in the text are replaced with their symbols according to their sentiment types. 
 
\subsection{Sentiment Word List Construction}
\label{swlc} 
As an alternative to existing approaches, we propose a new approach to extract opinion words and their sentiment strength from a labeled corpus using relationships between words and classes. In \cite{Dimitris'05}, positive and negative words are extracted from documents of
each class. In their algorithm, for each class $C$, all documents are examined and words are ranked based on their frequency in the document of the class. A set of top scoring positive words is labeled as good predictors of that class. A word $w$ is called positive for a class $c$ if \begin{equation}
P(c|w)>0.5*p+0.5*P(c)\label{eq1}
\end{equation} where $p$ is a parameter that is used in order to counteract the cases where the simpler relation $ P(c|w)>P(c)$ leads to a trivial acceptor/rejector, for too small/large values of $P(c)$.

 Similar to \cite{Dimitris'05}, we extract only positive words for each class.
We have both 10 class values as the sentiment strength of the text and 10 sentiment strength values for words from 1 to 5 and -1 to -5. After getting positive words of each class with using equation \ref{eq1}, we give class values to the selected words of that class as sentiment strength of the words. If the word $w_i$ is a positive word for class $i$, the value $i$ is given as its sentiment strength value. The intuition for this process is that if a word is seen more frequently in the text that has high sentiment strength, sentiment strength of that word should be also high and vice versa.

\subsection{Sentiment Strength Detection}

We propose two methods to measure the sentiment strength of tweets. As the basic method, we combine lexical and machine learning approaches, call it \textit{Feature selection using sentiment lexicon}. As the second method, a new feature type, called \textit{Opinion Based Text Representation: Grouped}, is created to represent the tweets. In addition to these, Sentistrength \cite{2} is configured for Turkish as an alternative to them.
\\ \newline
\textbf{Feature Selection Using Sentiment Lexicon}
 We use the lexicon to eliminate the words that do not affect the sentiment of the text. Just the words in the lexicon are kept and the others are removed. 
 Instead of looking only these words in the tweets as in the lexical approaches, we train a classifier using the words in the lexicon as the features.
 For this, we use the manually constructed and the automatically constructed lexicons as mentioned in the Section \ref{swlc}.\\ \newline
\textbf{Opinion Based Text Representation: Grouped}
BofW is the most commonly used method to represent the text as a feature vector. However, it constructs high dimensional feature vectors which include Each word in the documents a feature. However, two words that have the same sentiment strength would have the same effect on the overall strength of the text. This means that two synonymous words have the same influence on the sentiment strength of the text, so it does not matter which one is in the text. A feature can represent the presence of both of them. Just we need to know how many of them are seen in the text, since this may have an effect on the sentiment of the text. For instance if the number of the words whose sentiment values are 5 in a text is high, sentiment value of the text should be also close to 5. Therefore, the emotion words in the lexicon are grouped according to their sentiment strength. 

In our lexicon, there are five groups of positive and five groups of negative sentiment strength of the words. Also, one group for negation words such as ``de\u{g}il"(not), ``hay{\i}r"(no) and one group for booster words such as ``\c{c}ok" (much), ``fazla" (many) are added to them. In this new representation, each dimension of the feature vector corresponds to each group of the words. The last dimension of the feature vector is used to represent the sentiment strength (positive-negative) of the tweet as the class value.

The emotion words from groups are searched in the text. If a word from the group $i$ is seen in the text, the value of the corresponding dimension $i$ of the feature vector is raised by one. For instance, the word ``g\"{u}zel"(beautiful) has
strength of 3 as a positive sentiment. If a text includes this word, the third dimension of its feature vector is increased by one. As another example, the word ``i\c{s}kence"(torment) has strength of -4 value
as a negative sentiment. One is added to the value of the ninth (5+(-(-4)) dimension of the text's feature vector. Here is an example of the process of the converting
a text to a feature vector.

A tweet;

``Carrier-Z, your data plans are very good but your quality is zero.
I cannot speak well at home but I also cannot speak well outside" p:3
n:$-4$

Feature vector of the tweet; $<0,0,2,1,0,0,1,0,1,0,1,2,(2,4)>$

This shows that there are two words from group three and one word from group 4 and so on. After constructing feature vectors of tweets
according to the proposed representation, a classifier with one of machine learning algorithms is trained and used to find the sentiment of test data.

\section{Experimental Results}
\subsection{Data}

A collection of tweets about three different Turkish telecommunication brands gathered over one month is used as the corpus for our experiment. Tweets are judged on a 5 point scale as
follows for both positive and negative sentiment as in \cite{2} manually.
After eliminating Junk tweets, our data set includes 1420 tweets.

In addition to this, we construct a Turkish emotion word list manually as the lexicon. It includes 220 positive and negative words with a value from 1 to 5, following the format in
SentiStrength \cite{2} and booster words.

\subsection{Opinion Mining}
We tested our new feature type, \textit{Opinion based text
representation: Grouped}, and our proposed algorithms on the Turkish tweet data set by using a cross-validation approach.
The algorithms are called; with new representation \textit{Grouped}
with the word list constructed automatically \textit{Auto}
with the word list constructed manually \textit{Manual}. Also, the algorithms that include all words in the lexicons are called \textit{combin}.
We perform 10-fold cross-validation to train classifiers and to test
our new feature type and algorithm. Different machine learning
algorithms are used and best one with SVM is given in the table \ref{table:paps}  The results of our algorithms were compared to the result of the baseline majority
class classification, the classification obtained using BofW feature vectors type and SentiStrength configured for Turkish. For SentiStrength, we change its English word list with our manually constructed Turkish Word list.

When we change $p$ threshold values in the sentiment word list
construction algorithm, different words are selected. According to selected words, the results of classifications change. First, we perform experiments for different threshold values using different machine learning algorithms mentioned above and one of threshold values and machine learning methods that give the best result is selected for other tests. The best
threshold value is $0.4$ for negative sentiment strength and $0.8$ for
positive.

\begin{table}
\centering
\caption{Performance of algorithms on positive  sentiment strength
detection }\label{table:paps}
\begin{tabular}{|c|c|c|c|} 
\hline \multicolumn{2}{|c|} {Algorithms} & P. Accuracy & P. Accuracy $\pm
1$
\\ \hline \multicolumn{2}{|c|}{Baseline} & 74.54\% & 79.60\%
\\  \hline \multicolumn{2}{|c|}{BofW} & 57.45\% & 79.25\%  \\
 \hline \multicolumn{2}{|c|}{SentiStrength} & 56.90\% &73.17\%   \\ 
\hline 
\multirow{2}{*}{Combination} & Manual & 74.51\% & 82.78\%\\  \cline{2-4} & Auto & 55.70\%  & 74.54\% \\  \hline
\multirow{2}{*}{Grouped} & Manual & 75.50\% & 82.21\%  \\
 \cline{2-4} & Auto & \textbf{78.95}\%  &
\textbf{86.99}\% \\ \hline
\end{tabular}
\end{table}

\begin{table}
\centering
\caption{Performance of algorithms on negative sentiment strength
detection }\label{table:naps}
\begin{tabular}{|c|c|c|c|c|c|} 
\hline \multicolumn{2}{|c|} {Algorithms}&N. Accuracy & N. Accuracy $\pm1$
\\ \hline \multicolumn{2}{|c|}{Baseline}  & 43.10\% & 77.14\%
\\  \hline \multicolumn{2}{|c|}{BofW}   & 40.08\% & 73.14\% \\
 \hline \multicolumn{2}{|c|}{SentiStrength} & 30.77\% &
47.11\%  \\ 
\hline 
\multirow{2}{*}{Combination} & Manual & 48.42\% & 79.71\%\\  \cline{2-4} & Auto  & 56.27\%  & 81.59\%\\  \hline
\multirow{2}{*}{Grouped} & Manual  & 49.96\% & 81.17\%  \\
 \cline{2-4} & Auto & \textbf{62.94}\%  &
\textbf{84.32}\%\\ \hline
\end{tabular}
\end{table}

As we see from Table \ref{table:paps} and \ref{table:naps}, using BofW feature type is not sufficient for both positive and negative sentiment learning.
Since not all words affect the sentiment of the text and many
features make the learning difficult and give poorer results. After
selecting features using our lexicon, the results get better. However, after grouping them based on their sentiment strength, we
obtain the best results. Also, the lexicon automatically constructed gives better results than one manually constructed. Since it is constructed based on the dataset and sentiment strength of words may be different for different datasets. So, sentiments of words in the lexicon constructed manually may be different from the data set and this may not give good results for sentiment extraction.

\section{Conclusion}

In this paper, we present a framework for mining opinions by
extracting aspects of entities/topics on the collection of informal
short texts. We focus on Turkish and present a methodology
that can be applied to other languages where rich resources do not
exist.

We propose a new method to construct sentiment word list.
Then, we compute sentiment strengths of the text using novel feature
extraction algorithms. We apply feature selection by using sentiment
lexicons to reduce the complexity and improve the accuracy of the
results. Our framework includes methods to construct the sentiment
lexicons, and a novel opinion based representation of text that can
be applied to any language.


\bibliographystyle{plain} 
\bibliography{my}
\end{document}